Determining Sentencing Recommendations and Patentability Using a Machine Learning Trained Expert System

Logan Brown*
School of Engineering
University of Mount Union
1972 Clark Ave
Alliance, Ohio 44601
Phone: +1 (330) 823-3253
Email: brownlt2022@mountunion.edu

Reid Pezewski*
Department of Computer Science
University of Wisconsin-Milwaukee
3200 North Cramer Street
Milwaukee, WI 53211
Phone: +1 (414) 229-1122
Email: pezewsk7@uwm.edu

Jeremy Straub              *** Corresponding Author ***
Institute for Cyber Security Education and Research
North Dakota State University
1320 Albrecht Blvd., Room 258
Fargo, ND 58108
Phone: +1 (701) 231-8196
Fax: +1 (701) 231-8255
Email: jeremy.straub@ndsu.edu

* These authors contributed equally to this work

**Abstract**

This paper presents two studies that use a machine learning expert system (MLES). One focuses on a system to advise to United States federal judges for regarding consistent federal criminal sentencing, based on both the federal sentencing guidelines and offender characteristics. The other study aims to develop a system that could prospectively assist the U.S. Patent and Trademark Office automate their patentability assessment process.  Both studies use a machine learning-trained rule-fact expert system network to accept input variables for training and presentation and output a scaled variable that represents the system recommendation (e.g., the sentence length or the patentability assessment). This paper presents and compares the rule-fact networks that have been developed for these projects.  It explains the decision-making process underlying the structures used for both networks and the pre-processing of data that was needed and performed.  It also, through comparing the two systems, discusses how different methods can be used with the MLES system.

# 1. Introduction

A large collection of laws and regulations governs the operations of society.  In the United States, in most areas, individuals are regulated by federal laws (from the United States Code), federal agency rules (from the Code of Federal Regulations) and laws and regulations at the state and local level.  Ideally, these laws would be easily understood and applied to given circumstances.  Unfortunately, this isn't always the case.

In reality, laws and regulations are in conflict.  They are interpreted by the courts who, themselves, may provide conflicting interpretations, with some taking precedence over others.  Laws, agency regulations which implement them and prior court decisions must all be taken into account when attempting to determine what regulations apply.  This results in a regulatory patchwork which is, at best, confusing.

Computerizing legal interpretation is seen, by some, as a solution to this challenge.  The e-laws field utilizes data processing techniques to interpret and apply laws and regulations to real-world situations [1]. While there are numerous areas that e-laws could be implemented in, each has its own regulations and nuance.

This paper discusses the use of a machine learning artificial intelligence system, based on the gradient descent training of an expert system, for legal analysis related to patentability and criminal sentencing. These two areas were selected due to having existing well codified standards and their importance. Additionally, by applying the machine learning trained expert system (MLES) to these two different domains, its efficacy for e-laws project can be assessed, more generally.

Both application areas are poised to benefit from automation. Patent applicants will benefit from speed in protecting intellectual property. Automation can allow patents applications to be reviewed and acted on faster, providing patent provisions earlier.  For technologies that may become obsolete or which have a limited useful life, delays could mean that inventions go through much of their useful lifespan without patent protections.  In other cases, technologies won't be patented due to the length and cost of the process, as perceived by the inventor. Of course, patents should be decided upon consistently and fairly.

Criminal defendants also need to be ensured that their sentences are decided fairly, without prejudice or bias of any type.  While current sentencing guidelines are designed to promote this, interpretation and discretion still exists.  Due to this, judges may unknowingly give harsher penalties to different groups without the intent to do so [2].  Differences in interpretation, even without bias, could also result in significant unintentional sentencing differences.

The purposes of the two studies presented in this paper are varied; however, both seek to remedy current issues.  The slowness of the patent application process prospectively harms the intellectual property benefits that could be enjoyed by applicants [3].  Similarly, the alleged biases in criminal sentencing [2] lead to some paying a harsher penalty for similar crimes than others. The numerous applications of machine learning for judicial and quasi-judicial uses presented in [1] suggest that that machine learning may be effective in aiding decision makers by suggesting outcomes (based on predicting what a human decision maker would have decided) and preventing inconsistencies in decision-making.

This paper presents both studies and explains the need for the use of machine learning in both application areas.  In each case, the datasets that are used and how they are used by the MLES system will be discussed. The structure implemented for the MLES rule-fact network, developed for each application area, will also be explained with a focus on describing why the rule-fact network is structured in the particular way that was chosen.  The relationships between components of the rule-fact network will be described.  The paper will also compare the use of the MLES system for both areas before concluding and discussing potential areas of future work.

## 2. Background

This section reviews prior work in several areas relevant to the current study. First, prior work on artificial intelligence is reviewed. Then, work on expert systems is considered. Finally, a review of prior work on gradient descent training for expert systems is provided.

### 2.1. Artificial Intelligence

Artificial intelligence (AI) is a key technology that has found numerous uses in modern society. AI systems have been demonstrated to provide benefits [4] including find software bugs [5], helping handicapped individuals [6], detecting cyberattackers [7] and controlling robots [8]. Some AI systems have a capability to learn on their own [9] or the ability to learn using a well-defined training process. Examples of learning approaches include the use of rewards and supplied input and output data [10]

One form of AI training, gradient descent and backpropagation [11], is integral to the work presented herein. This technique applies changes, typically to a neural network, based on an iterative difference correction process. Multiple forms of gradient descent have been proposed. Key among these are techniques that seek to eliminate bias [12], [13] or increase speed [14], allowing more training to occur within a period of time, potentially leading to higher quality decisions or recommendations. Other enhancements have focused on and resilience to AI-targeting attacks [15], [16]. Notably, other AI techniques including genetic algorithms [17] and particle swarm optimization [18] have also been proposed for use in neural network training. The system utilized herein applies gradient descent to an expert system.

### 2.2. Expert Systems

Expert systems are also a form of AI. Classical expert systems use rule-fact networks and inference engines [19] to make recommendations or decisions. Systems, which were designed to replicate the expertise of a human expert, were first introduced in the 1960s and 1970s [20], [21].

The most basic expert systems have Boolean facts and rules that are logical 'AND' operators. A rule processing engine scans the rules and facts to identify rules whose input facts are true and runs them, setting an output fact (or facts) to true as well. More advanced expert systems which use fuzzy logic concepts [22] have also been proposed, among several other types. Fuzzy logic systems can store data regarding uncertainty or partial membership in facts and have more nuanced rules. Other forms of expert systems have been developed which use other AI techniques such as genetic algorithms and neural networks [23].

Prior to the work presented in [24], which serves as the core system used for this work, Mitra and Pal [25] defined a "knowledge-based connectionist expert system" which started with "crude rules" that were stored as neural network connection weights and optimized. Problematically, this system does not appear to have been implemented. It is also unclear how training away network characteristics would be prevented by this type of a system.

Expert systems have found wide use in areas such as agriculture [26], power systems [27] and medicine [28], [29]. Systems based on fuzzy logic have been demonstrated for medical applications [30]–[32], software architecture evaluation [33] and autonomous vehicle control [34].

### 2.3. Gradient Descent Expert Systems

The work presented herein utilizes a machine learning-trained expert system which was presented in [24]. This system is designed to be defensible, meaning that it (and the decisions that it makes) are human understandable and that it has protections against learning invalid (non-causal) relationships. The system combines the rule-fact network of an expert system and the training capabilities of a neural network.

The system performs gradient descent directly on the expert system rule-fact network. Thus, while some conceptual similarity to using a neural network for expert system rule development and refinement exists [25], the approach proposed in provides benefits beyond this by not allowing network-changing learning. Additionally, [24] showed that the training process (which distributes error-correction to rules' input fact weightings) can be very expedient with a well-defined expert system network.

The system utilizes fact values between 0 and 1, thus supporting the concepts of partial membership and ambiguity. Rules weight the impact of the input facts with regards to the output fact, instead of being an 'AND' operator. Each rule, in the system, has two input and one output facts. A weighting is assigned to each input fact, which must be between 0 and 1; the two weightings for a rule must sum to 1. A specialized training algorithm, to support expert system rule-fact networks, was proposed in [24].

Figure 1 shows how the system works by depicting its overall operations. The level of change determination algorithm is shown in Figure 2.

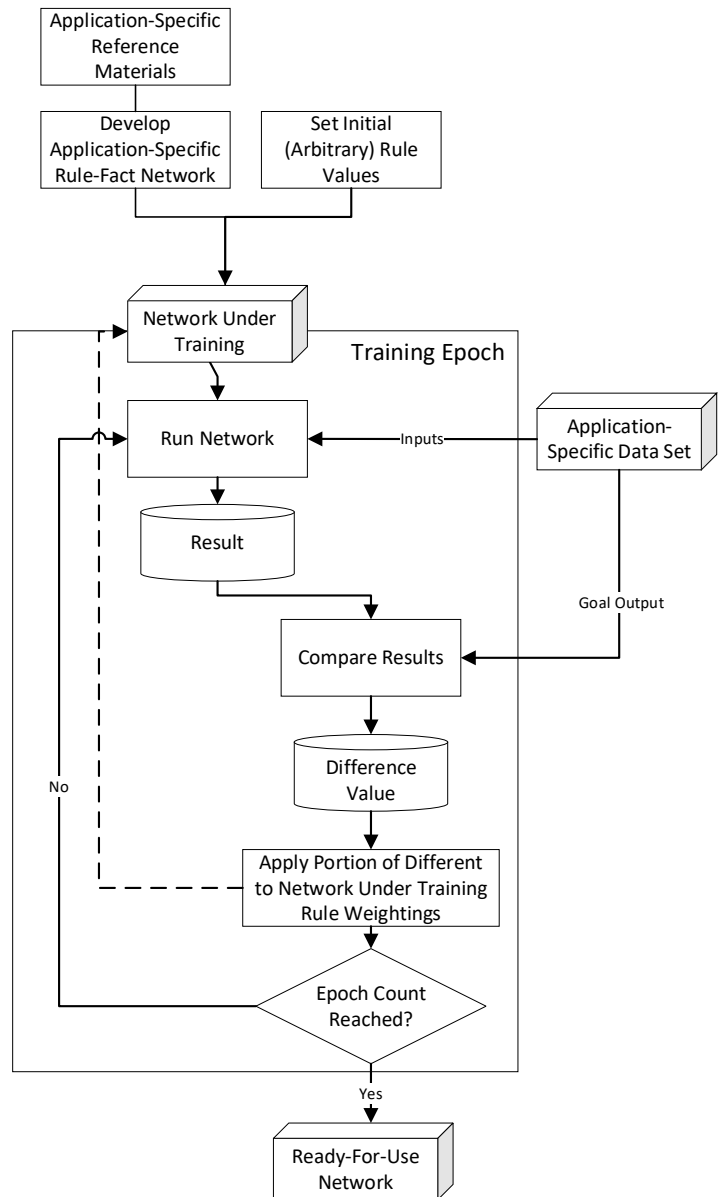

Figure 1. Training Process (based on [24]).

Fundamentally, the system operates by identifying the rules that contribute to the identified output fact. The results produced by the system being trained are compared to data from the training dataset, during the training process. If a difference between the two exists, a portion of it is applied as a correction to each identified contributing rule. The amount of change that is made is determined by a user-configurable velocity value. While the system was demonstrated in [24] using a 'perfect' system training technique (described in [35]), for these studies datasets are used as training inputs.

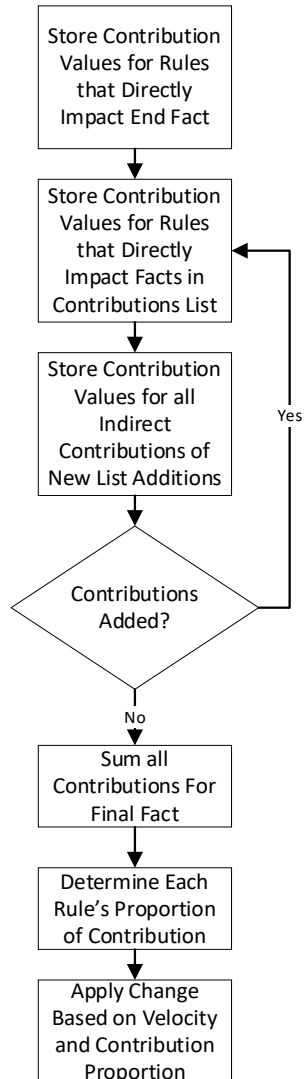

Figure 2. Algorithm for determining node change levels [24].

## 3. Sentencing Recommendation Study

The sentencing recommendation study is assessing the efficacy of using the MLES for making recommendations based on offender crimes, characteristics and the USSC Guidelines [36]. The goal is to develop a system that makes consistent recommendations, based on relevant guidelines, that approximate what judges, collectively, would sentence an offender to (if all non-relevant offender characteristics were ignored). The use of the MLES is designed to prevent system bias while the use of a computational system is designed to combat the potential for human bias. Concerns related to each are now discussed.

### 3.1. Issues with existing machine learning systems and human bias in the criminal judicial system

Machine learning systems have previously been used and are currently in use within the United States legal system for sentencing purposes. An Arnold Foundation algorithm and software package was widely used at one point -- twenty-one jurisdictions referred to it as part of criminal proceedings [37]. The algorithm was used in Florida to determine if a defendant was eligible for bail and, if so, to set the bail amount [37]. However, allegations that the system was biased were made [38]. It was asserted that the system's consideration of defendants' socio-economic positions gave certain groups a disadvantage within the system, resulting in the system unintentionally acting in a discriminatory manner [38].

However, computer systems are not the only bias in the legal system. Research [1], which incidentally used machine learning technologies for analysis, was performed in California which looked at the parole decision making process. Based on the analysis of thousands of transcripts from hearings, keywords that have the largest effect on the outcome of the hearing were identified. The keywords identified included, as would be expected, the offender's previous criminal history, the type of offense, the number of victims, and how the offender had spent their time in prison. Prisoners using their time for remorseful purposes, to plan their future, to increase their education, and who overall have a less pronounced criminal history were more likely to be released. This is to be expected and is based on clearly relevant factors. Problematically, the study indicated that other factors also affected the outcome. Cases requiring the use of an interpreter, for example, had a 7% reduced rate of being given parole. Cases where a psychologist evaluated an inmate resulted in parole award reductions of between 3-7%. This would suggest that offenders were, effectively, being punished for not speaking English and for certain types of disabilities [1]. Prospectively, a well-developed computational tool could remove or reduce these types of inconsistencies within the criminal justice system.

### 3.2. Project goals

The goal of the sentencing recommendation study is to assess the efficacy of an MLES-based system to provide recommendations to judges on offenders' sentence length. The implementation focuses on two main components: offense characteristics and the offender's criminal history.

### 3.3. Data, analysis, and processing

A dataset was obtained from the United States Sentencing Commission (USSC) website [39]. The dataset contains over 100 variables relevant to sentencing procedure, however not all of them are useful in determining the sentence length. Twenty-one of these variables were identified for use as input facts to the network. These facts were identified based on their close alignment with the guidelines in the USSC Guidelines Manual [36]. The facts all contain of the information pertinent to calculating the final offense level and the final criminal history level, as defined by the USSC sentencing guidelines.

These variables contain the base offense level (BOL), special offense characteristics (SOC), Chapter 3 adjustments (which are characteristics that can be applied to any crime), and Chapter 4 criminal history and livelihood variables. The most important variables in determining offenders' sentence length are the BOL, SOCs, and criminal history characteristics like the POINT1, POINT2 and POINT3 fields, which describe offender's prior incarceration duration. The meaning of every variable and their values are detailed in the USSC codebook [40], which has been publicly released.

As the MLES uses data on a scale between 0 and 1, the variables from the dataset were converted to this scale. Variables that have a binary value, were simply set to 1 for applied and 0 for not applied. Other values used a scaling factor to place them within the 0-1 range: they were simply divided by the maximum relevant value. This approach supported both well-defined variables as well as those, such as the BOL and SOC, where there is not a set limit for the maximum in the guidelines.

At the point that sentencing is occurring, the defendant's guilt has already been determined. The offense characteristics are, thus, known and are separated into different variables that are added together using a point-like system. This process starts with the BOL corresponding to each crime, as listed in the sentencing guidelines manual. The SOCs relevant to the crime and afterwards are then added. Next, adjustments based on Chapter 3 of the sentencing guidelines manual (these are adjustments that increase the offense level based on characteristics such as the role of the perpetrator, the status and effect on the victims and other crime agnostic characteristics) are made.

The criminal history is a set of variables which focus on the offender's previous crimes. These variables contain information on how many offenses were previously committed with a sentence of under 60 days (POINT1), with sentences between 60 days and 13 months (POINT2) and sentences longer than 13 months (POINT3). Crimes committed shortly after being released also increase the criminal history level. The RLEASHI variable indicates whether an offender falls under this rapid reoffence category. Using all of this data, which are inputted as facts into the MLES-based system, the system is able to compare the case to others with similar characteristics to determine a sentence recommendation for the defendant.

The sentencing recommendation system utilizes the point values given from the sentencing variables, which are pre-processed to be on a 0 to 1 scale, to calculate an output value which is a recommended sentence duration. This is output as a value with a 0 to 1 range as well, which is converted back to a number of years and months. Notably, 0 indicates that no prison sentence should be given and 1 indicates a recommended life sentence.

### *3.4. Rule-fact network design*

The rule fact-network, implemented in the MLES, is discussed in this section and depicted in Figures 1 to 5. Figure 1 provides an overview of the structure of the network, while Figures 2 to 5 provide detail views of each area. As shown in Figure 1, the facts are separated into the two main branches: offense level characteristics and criminal history characteristics. Within these branches, the facts are grouped based on other characteristics similarity. For example, the role adjustments are combined together while the criminal history points are combined.

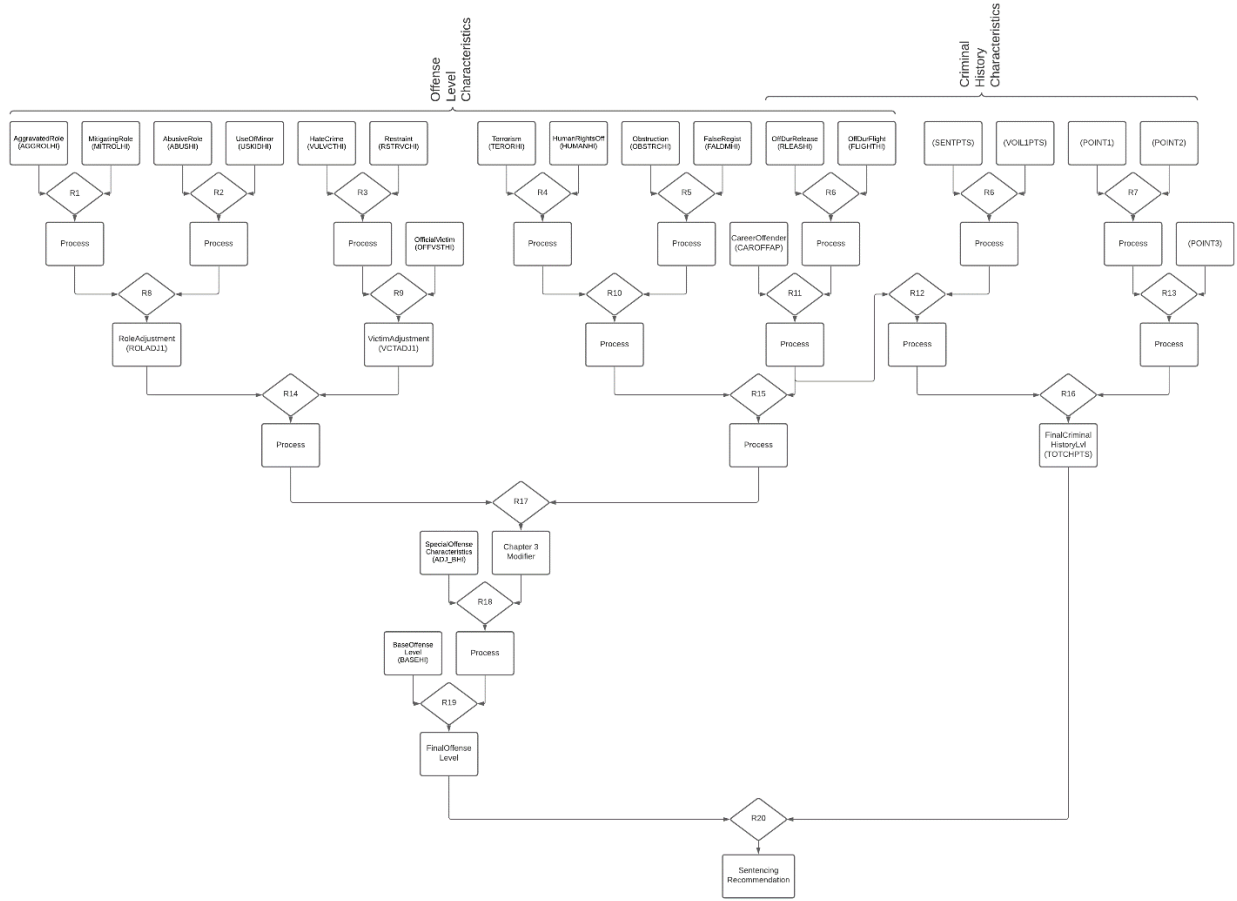

Figure 1. Sentencing study rule-fact network overview.

The Chapter 3 adjustments, which are related to the role of the perpetrator, and the role of the victim, are depicted in Figure 2. This area of the network includes factors such as whether the crime had aggravating, mitigating or abusive roles and whether it involved the use of a minor. This are of the network also includes victim characteristic-based adjustments such as if a crime is determined to be a hate crime, the use of restraints, and the official victim (i.e., whether the offender misused an official position in the crime) status. Note that facts labeled 'process' in the diagrams represent intermediate facts which are used to facilitate the use of combinations of rules for combining data from multiple facts.

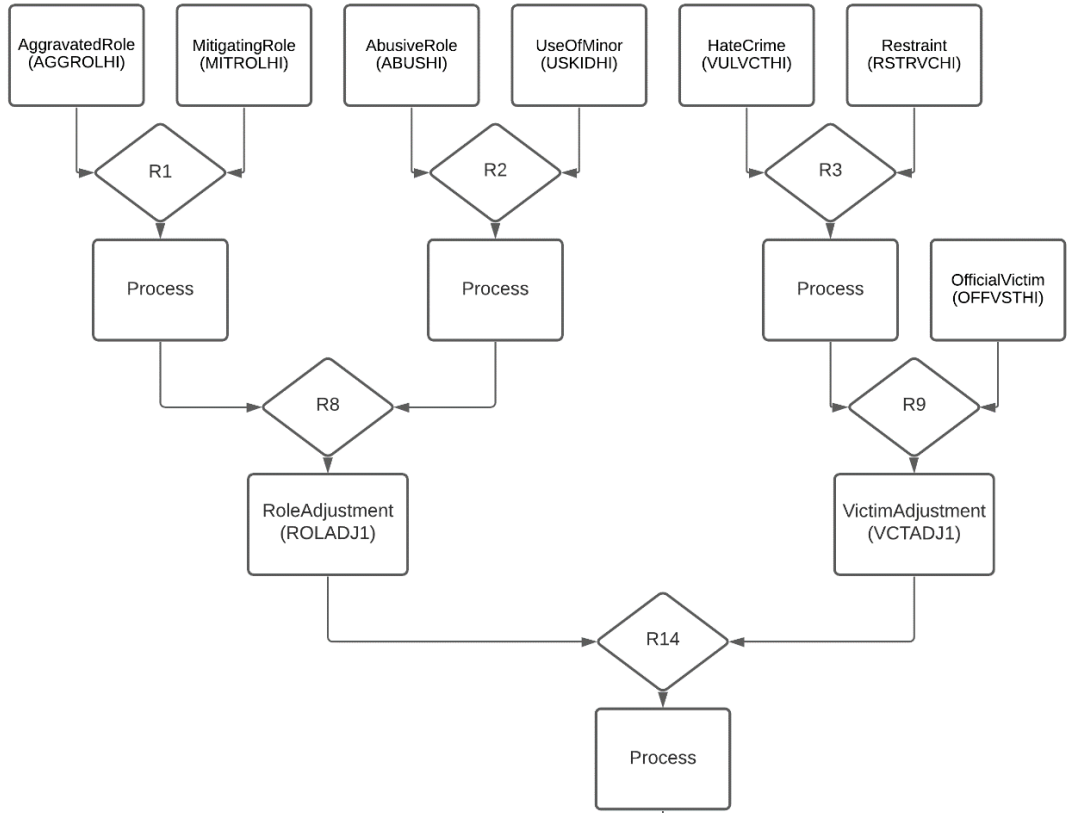

Figure 2. Chapter 3 adjustments portion of the sentencing system rule-fact network.

Figure 3 shows additional Chapter 3 adjustments including whether the crime was a form of terrorism, whether it constituted a human rights offense, whether obstruction of justice occurred and whether it was based on false domain name registration. These are all depicted on the left side of the figure. The right side of the figure shows three facts that are considered in both the offense level and criminal history categories: whether the defendant is a career offender, whether offenses were committed during release before sentencing, and whether offenses were committed during the defendant's flight from law enforcement.

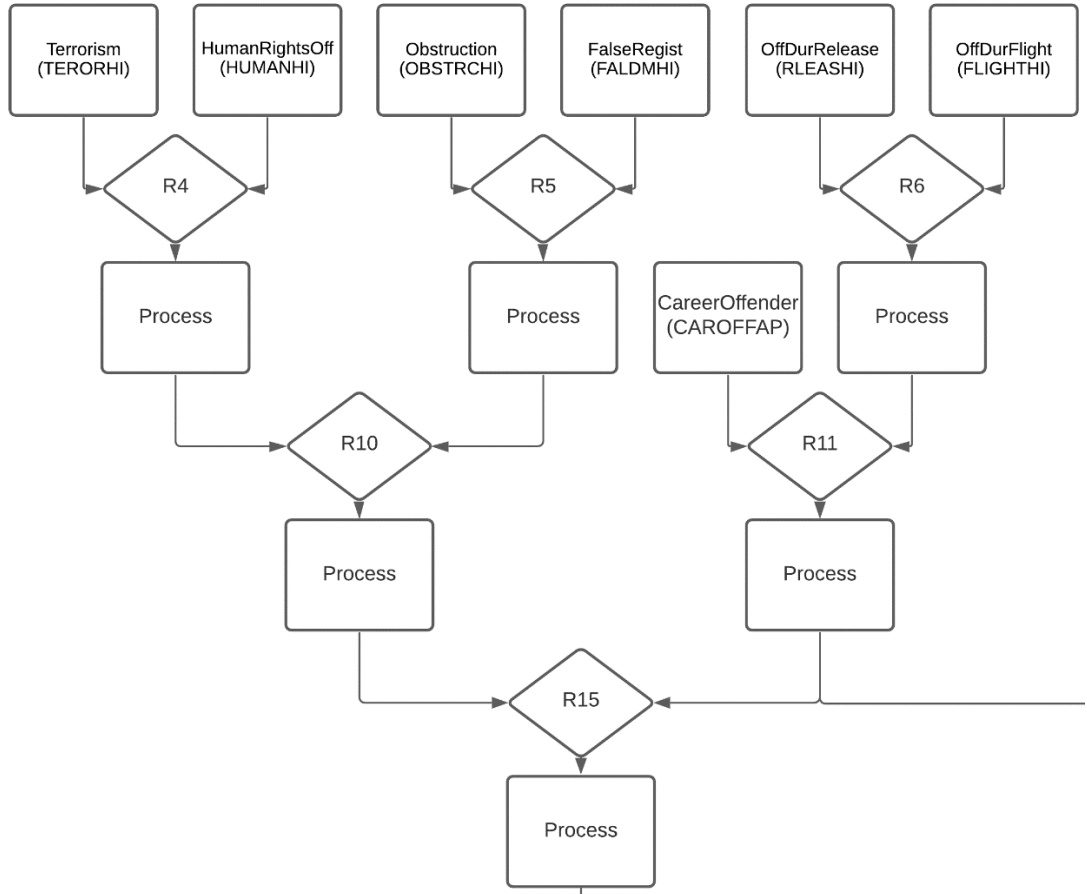

Figure 3. Chapter 3 and 4 adjustments portion of the sentencing system rule-fact network.

The main facts used in determining criminal history level, including the POINT1, POINT2, and POINT 3, variables and the SENTPTS variable (indicating if a criminal committed an instant offense while the current case was ongoing) and the VOIL1PTS variable (which includes any other criminal activity not included in other facts) are depicted in Figure 4. Note that weightings are used to divide the range of possible sentence values to give significantly more weight to the higher impact sentences. Rule 12 connects these variables to the other criminal history variables depicted in Figure 3. Rule 16 outputs the total criminal history points value, which is used in determining the sentence length, as depicted in Figure 5.

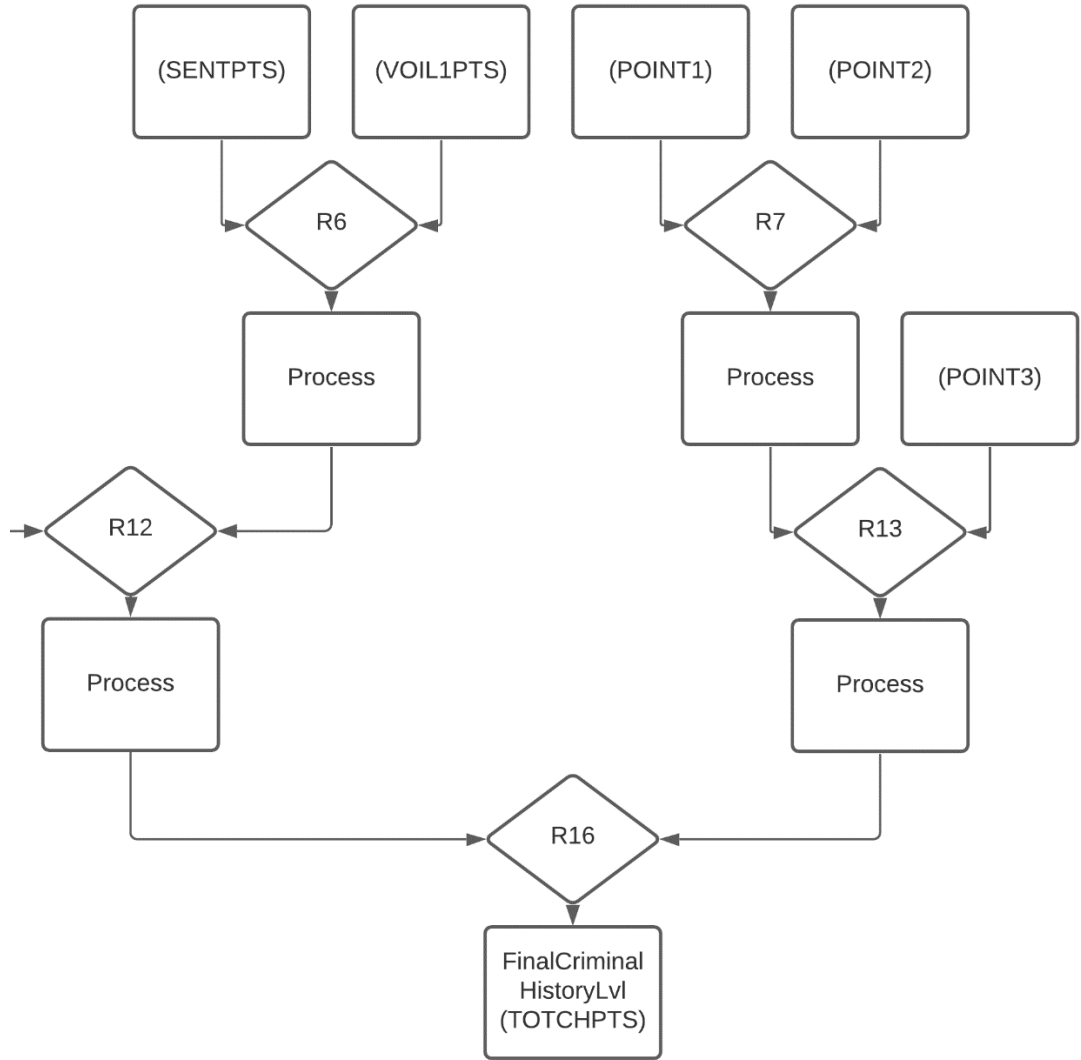

Figure 4. Criminal history portion of the sentencing recommendation system rule-fact network.

Finally, the remainder of the sentencing system rule-fact network, shown in Figure 5, brings all the variables together to recommend a sentence. Rules 14 and 15, which are depicted in both this figure and Figures 2 and 3, respectively, bring the Chapter 3 and Chapter 4 modifiers. This modifier is them merged with the SOC's and the BOL to calculate the final offense level. Rule 20 calculates the recommended sentence length by combining the final offense level, and the final criminal history (from Figure 4).

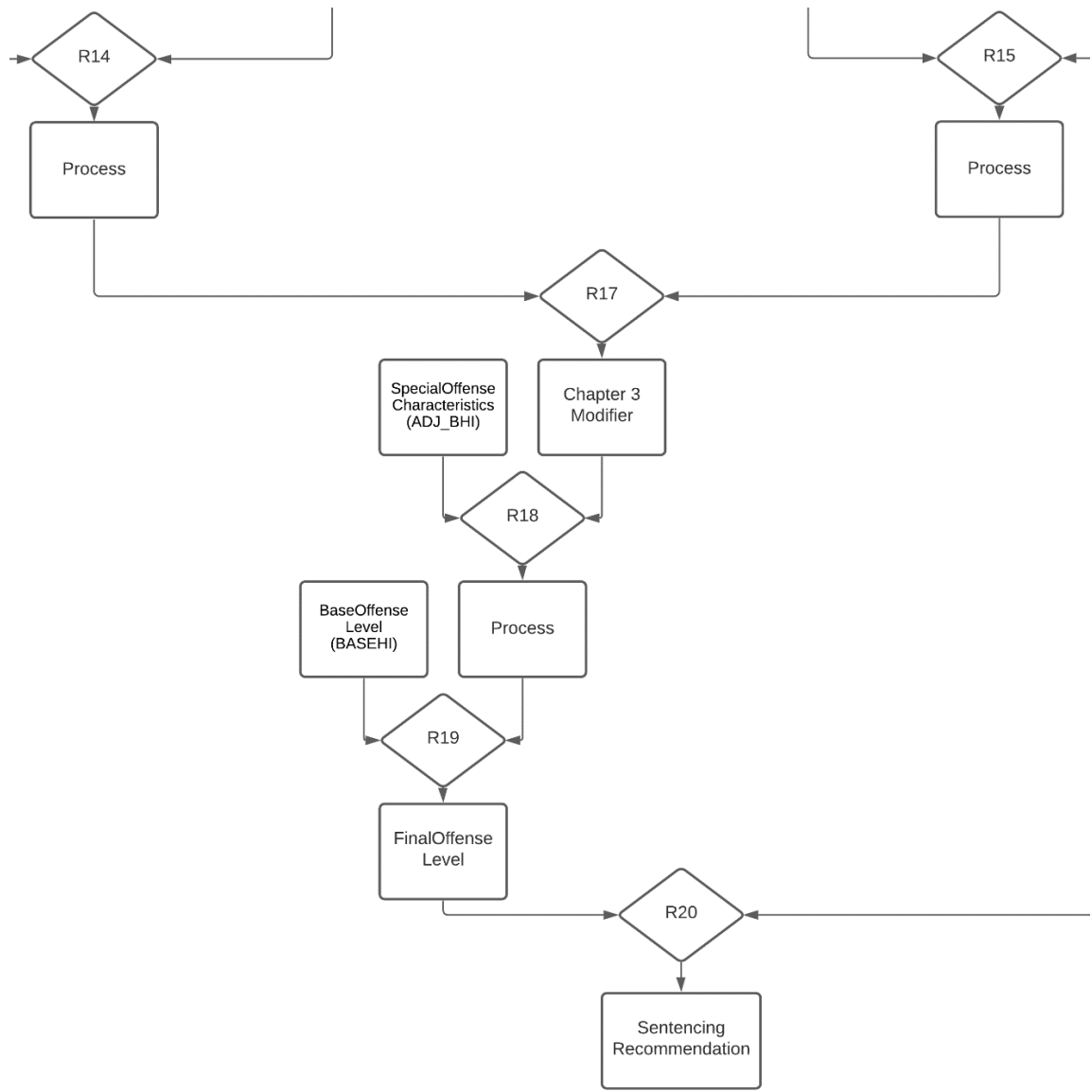

Figure 5. Final portion of the sentencing system's rule-fact network.

## 4. Autonomous patentability determination study

There is limited prior work which has used machine learning for assessing patentability. Fortunately, the U.S. Patent and Trademark Office (USPTO) has made several datasets available which can be used to answer parts of this question. This study, for example, uses a 2021 dataset from Giczy, Pairolero and Toole [41] which was designed to facilitate answering questions regarding the patenting of artificial intelligence-related technologies.

The patentability study focuses on the patentability requirements outlined in §102 and §103 of the United States Code (USC). These sections require, at their most basic level, patentable technologies to be unique and unable to be easily produced by the public [42], [43]. The USC, the Code of Federal Regulations (CFR), and USPTO procedures define numerous other requirements as well as procedural steps that must be followed.

Patent examiners analyze the facts presented in the patent application about the novelty and utility of a technology. They must also search for so-called prior art – instances of the same technology or similar technologies which limit the novelty of the technology that a patent is sought for. Based on their analysis, they must make a decision regarding the denial or award of the patent. This process is inherently iterative, with patent examiners finding prospective issues that may prevent the issuance or limit a patent and the inventor (or their attorney) seeking to overcome these challenges while maintaining the broadest scope of covered technology as possible.

The patent examiner's analysis inherently has room for error, as a human-performed process. Additionally, the examiner and attorney's more limited knowledge of the invention, as compared to the inventor, may result in both erroneous award and denial decisions. Even after the extended patent examination process, prior art which was unknown to the examiner may come to light, raising questions about patent issuance.

Given all of this, the goal of this system is not to replicate or automate the necessarily very bespoke analysis that patent examiners must perform. Rather, it is to try to determine what patent applications are most likely to succeed, to allow inventor, attorney and examiner time and resources to be focused on the most beneficial areas possible.

The process used by patent examiners to assess a patent application, embodied into a MLES-style rule-fact network is presented in Figure 6. To assess the patent application, prior art must be identified and compared to the invention to determine its level of difference from existing technologies. Additionally, the patent examiner must determine whether any public disclosures of the technology have occurred which may prevent or provide timeframe limitations patentability. Key input facts to the MLES system include data regarding prior commercial use, sales, publications, public presentations, prior public use, materials contained in unpublished patent applications and foreign patents and applications. Additionally, the examiner must determine how easily the technology could be developed by a member of the public (without the knowledge contained in the patent application).

This project is in its early phases and the basic rule-fact network for its operations has been developed. Future work is needed to answer many of questions that these needed inputs pose. Whether this data can be arrived at directly or approximated is a key question being answered by ongoing work.

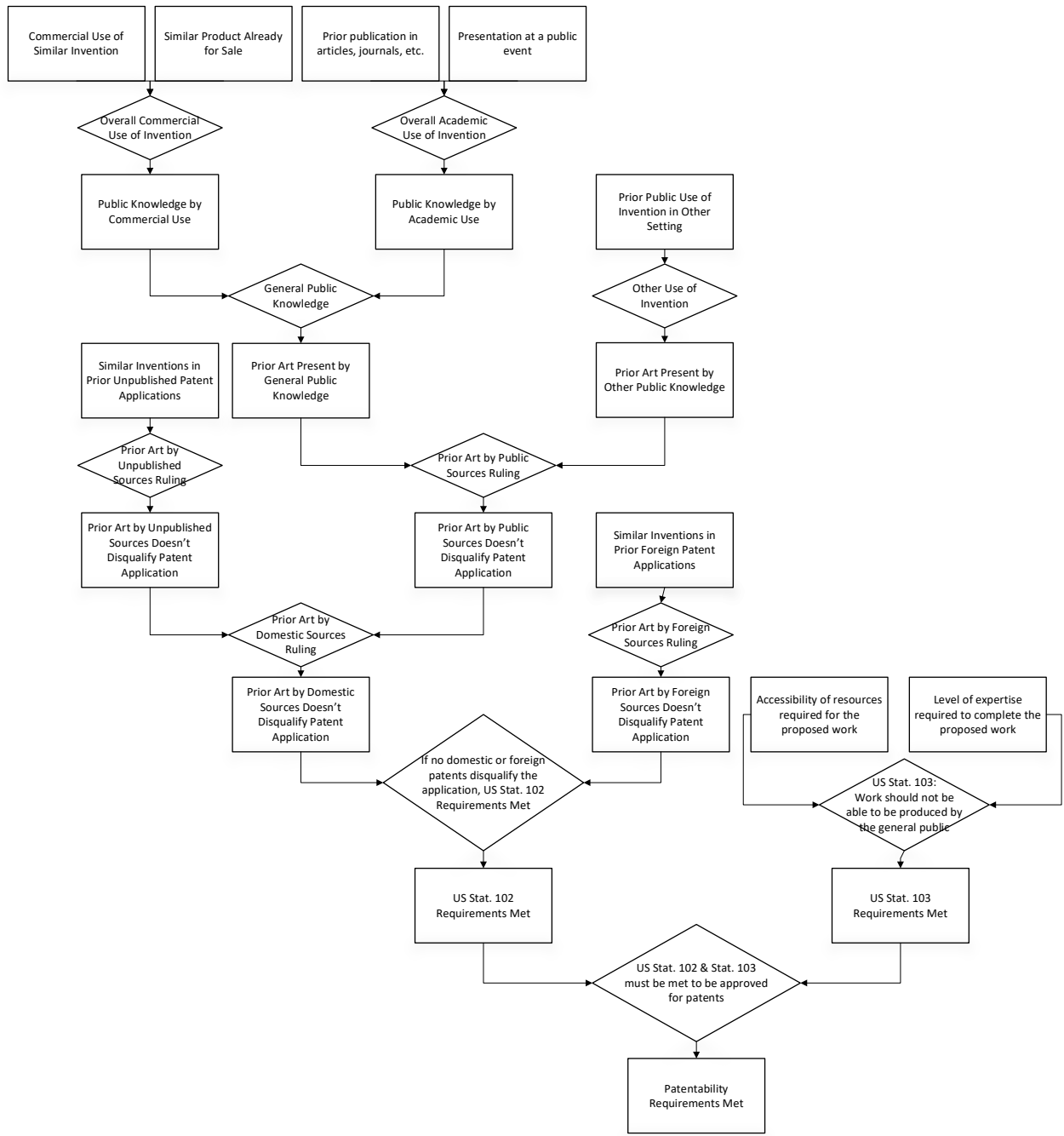

Figure 6. Patentability analysis rule-fact network for use with MLES.

The patentability assessment MLES system rule-fact network is based on the USPTO patentability requirements as indicated in the Patent Examiner's Handbook's [44]. As with the sentencing recommendation rule-fact network, it starts with presumably obtainable facts and combines these values leading to a result as to how likely a patent is to be issued, ranging from 0 being the least likely to be approved and 1 being the most likely.

The specific fact nodes that are included in the network have been determined based on an analysis of data gathered from the US Patent and Trademark Office's Patent Examiner's Handbook and its outline of what defines patentability [44]. The system combines a decision-making process related to determining if prior art exists, from a strict legal definition perspective. If the presence of prior art is identified, then the difference from this prior art must be assessed to determine if it is significant to merit patentability. Other factors, such as the ability of members of the public to develop the invention on their own, also form part of the assessment.

Within the rule-fact network, there are four primary branches: domestic public prior art, domestic private prior art, foreign prior art, and what is needed to develop the invention. Each branch funnels down into the final patentability review. This follows the standards set by USC §102 and §103 to determine if a patent application meets the requirements for securing intellectual property protections.

Fundamentally, the data that will be used to train and assess the proposed system comes from the USPTO. The exact method of accessing this data remains an area of ongoing work. A vast amount of data is available via the USPTO Public PAIR system. Additionally, multiple datasets are available that have been previously created for different purposes; however, these may not have all of the data in the Public PAIR system. Multiple pieces of needed data were identified. Of course, one of the most important is the status of the patent (whether it was issued, is in progress, or was abandoned).

Some data that is in Public PAIR may need to be converted from image files to computer-readable text and processed with text mining or other tools to generate the requisite inputs to the MLES-based system and the rule-fact network described previously.

## 5. Comparison of Techniques Used

While both studies were designed based on using the same MLES system to implement their rule-fact network, their networks are somewhat different. The patentability study effectively combines what could be a mix of largely Boolean inputs that the system would use to make a decision on, but it is probable that there is more nuance – as a small deficiency in one area may be offset by other areas. For example, the novelty of a technology and the ease of its creation would have an inherent balancing between them. This, however, is an assumption that remains to be tested. Similarly, while the existence of prior art could perhaps be taken to be a Boolean input, the reality is that this is nuanced judgement call which can (and for prediction accuracy, likely must) be approximated (until the matter is definitively resolved).

While these values can be predicted when attempting to anticipate the result of patent adjudication, they are reduced – largely – to Boolean variables, if the system was to be used to assist a patent examiner who was making the decisions her or himself. However, the system could, prospectively, utilize projected values (which would later be replaced by real ones) to facilitate prioritization and other decision making throughout the patent adjudication process, until each category was decided by an examiner. Additionally, discrepancies between the examiner decision and the system decision could, prospectively, trigger a quality assurance review.

The network design utilized by the sentencing system, conversely, has a more variable set of inputs. A limited number are Boolean-based or have fixed values; however, many others support a continuum of values. Additionally, as there are numerous inputs that determine the result of each system area, the intermediate facts can store and convey considerable nuance. From these facts and the rules that process them, a value of between 0 and 1 results that can be scaled to provide the system's recommendation for the number of months for the offender to serve in prison.

The network designs are also noticeably different. The patentability study largely follows a linear approach, where conditions could prospectively be checked in a given order. While this might be ideal, if a single examiner was reviewing a patent application – as it would allow easier-to-assess potential disqualifying conditions to be assessed prior to more difficult-to-assess ones – the approximation values used by the MLES system should be able to be concurrently loaded, in most cases. This is demonstrably different from the multiple coalescing branches of the sentencing study's network, which stay separate until combination by the final rule.

The systems differ in their target users and how they are used by them, though similarities exist. The sentencing study aims to create a tool to help judges make – and prosecutors and defense attorneys predict – federal sentencing outcomes. The patentability assessment system, similarly, helps the patent examiner – who has a similar adjudicating role (albeit one that would eventually be reviewed by a judge, should the USPTO and applicant disagree as to the assessment outcome) – predict a patent allowance determination. Thus, while the sentencing study aims to produce a process aiding tool, the patentability study, instead, seeks to help patent applicants predict the outcome of an application and examiners (and their managers) optimize workflow and conduct quality assurance.

Although the two projects have different purposes and differences in execution, the logic is similar and able to be modeled by the MLES system. While training, each processes facts about a case (application) and uses them to update stored knowledge (in the network weights). When a case (application) is presented, these stored weightings, as well as the network structure itself, are used to make a prediction or recommendation. Each system must, similarly, be trained using data from thousands of cases (applications).

## 6. Conclusion

The two studies should determine whether the MLES system can be used to aid and streamline the processes of criminal defendant sentencing and patent application review, while simultaneously removing biases that may be present from humans deciding the outcomes unaided. Initially, neither project is intended to replace human decision-making. Instead, the systems are designed to help humans to make better decisions more efficiently.

The current judicial system has been alleged to have biases in defendant sentencing. The sentencing guidelines, themselves, were an attempt to create more uniformity in the sentences given defendants that had committed similar crimes under similar circumstances. However, differences in circumstances, interpretations and judicial discretion leave room for inconsistencies and prospective bias.

This implementation of the sentencing guidelines within the MLES system, following the USSC's guidelines and using USSC data for training may help to identify key discrepancies in sentencing as well as to increase the fairness, efficiency and predictability of sentences. Similarly, the use of the patentability assessment system may facilitate applicants and examiners devoting time and attention to the most promising technologies' patent applications. This will save applicants of likely-to-be-rejected applications money and time while allowing successful applicants to, potentially, benefit from intellectual property protection earlier.

In both cases, work is ongoing to assess the efficacy of the MLES system as well as the individual networks that have been developed. Network adaptation may be necessitated after initial studies are completed. Additionally, the utility of the system to its target user base remains to be assessed.

**Acknowledgements**

This work has been partially supported by the U.S. National Science Foundation (NSF Award # 1757659) and the NDSU Challey Institute for Global Innovation and Growth. Facilities and other resources have been provided by the NDSU Department of Computer Science. Thanks is given to Benjamin Fitzpatrick

## References

[1] M. A. Livermore and D. N. Rockmore, *Law as Data: Computation, Text, and the Future of Legal Analysis*. Sante Fe Institute Press, 2019.

[2] L. Ware, T. Branch, J. Lewis, L. F. Litwack, M. L. Oliver, and T. M. Shapiro, "Leland Ware, Louis L. Redding Professor University of Delaware. This presentation is derived from," vol. 68, no. 2006, pp. 1–15.

[3] S. Totten, "Local inventors frustrated over patent office bureaucracy," *2009*, Minneapolis, Oct. .

[4] M. Nadimpalli, "Artificial Intelligence Risks and Benefits," *Int. J. Innov. Res. Sci. Eng. Technol.*, vol. 3297, 2007, Accessed: Jun. 25, 2020. [Online]. Available: https://www.researchgate.net/publication/319321806.

[5] A. Tosun, A. Bener, and R. Kale, "AI-Based Software Defect Predictors: Applications and Benefits in a Case Study," Jul. 2010. Accessed: Jun. 25, 2020. [Online]. Available: https://www.aaai.org/ocs/index.php/IAAI/IAAI10/paper/view/1561.

[6] H. A. Yanco and J. Gips, "Preliminary investigation of a Semi-Autonomous Robotic Wheelchair Directed Through Electrodes," in *Proceedings of the Rehabilitation Engineering Society of North America Annual Conference*, Jun. 1997, pp. 414–416, Accessed: Jun. 25, 2020. [Online]. Available: www.cs.bc.edu/~gips/EagleEyes.

[7] Z. A. Baig, M. Baqer, and A. I. Khan, "A pattern recognition scheme for Distributed Denial of Service (DDoS) attacks in wireless sensor networks," in *Proceedings - International Conference on Pattern Recognition*, 2006, vol. 3, pp. 1050–1054, doi: 10.1109/ICPR.2006.147.

[8] S. C. Jacobsen *et al.*, "Research Robots for Applications in Artificial Intelligence, Teleoperation and Entertainment," *Int. J. Rob. Res.*, vol. 23, no. 4–5, pp. 319–330, Apr. 2004, doi: 10.1177/0278364904042198.

[9] G. Paliouras, C. Papatheodorou, V. Karkaletsis, and C. . Spyropoulos, "Discovering user communities on the Internet using unsupervised machine learning techniques," *Interact. Comput.*, vol. 14, no. 6, pp. 761–791, Dec. 2002, doi: 10.1016/S0953-5438(02)00015-2.

[10] R. Caruana and A. Niculescu-Mizil, "An empirical comparison of supervised learning algorithms," in *ACM International Conference Proceeding Series*, 2006, vol. 148, pp. 161–168, doi: 10.1145/1143844.1143865.

[11] R. Rojas, "The Backpropagation Algorithm," in *Neural Networks*, Berlin: Springer Berlin Heidelberg, 1996, pp. 149–182.

[12] C. Aicher, N. J. Foti, and E. B. Fox, "Adaptively Truncating Backpropagation Through Time to Control Gradient Bias," in *Proceedings of The 35th Uncertainty in Artificial Intelligence Conference*, Aug. 2020, pp. 799–808, Accessed: Feb. 22, 2021. [Online]. Available: http://proceedings.mlr.press/v115/aicher20a.html.

[13] L. Chizat and F. Bach, "Implicit Bias of Gradient Descent for Wide Two-layer Neural Networks Trained with the Logistic Loss," *Proc. Mach. Learn. Res.*, vol. 125, pp. 1–34, Jul. 2020, Accessed: Feb. 22, 2021. [Online]. Available: http://proceedings.mlr.press/v125/chizat20a.html.

[14] R. Battiti, "Accelerated Backpropagation Learning: Two Optimization Methods," *Complex Syst.*, vol. 3, no. 4, pp. 331–342, 1989, Accessed: Feb. 22, 2021. [Online]. Available: https://www.complex-systems.com/abstracts/v03_i04_a02/.

[15] P. Zhao, P. Y. Chen, S. Wang, and X. Lin, "Towards query-efficient black-box adversary with zeroth-order natural gradient descent," *arXiv*, vol. 34, no. 04. arXiv, pp. 6909–6916, Feb. 18, 2020, doi: 10.1609/aaai.v34i04.6173.

[16] Z. Wu, Q. Ling, T. Chen, and G. B. Giannakis, "Federated Variance-Reduced Stochastic Gradient Descent with Robustness to Byzantine Attacks," *IEEE Trans. Signal Process.*, vol. 68, pp. 4583–4596, 2020, doi: 10.1109/TSP.2020.3012952.

[17]   J. N. D. Gupta and R. S. Sexton, "Comparing backpropagation with a genetic algorithm for neural network training," *Omega*, vol. 27, no. 6, pp. 679–684, Dec. 1999, doi: 10.1016/S0305-0483(99)00027-4.

[18]   A. Saffaran, M. Azadi Moghaddam, and F. Kolahan, "Optimization of backpropagation neural network-based models in EDM process using particle swarm optimization and simulated annealing algorithms," *J. Brazilian Soc. Mech. Sci. Eng.*, vol. 42, no. 1, p. 73, Jan. 2020, doi: 10.1007/s40430-019-2149-1.

[19]   D. Waterman, *A guide to expert systems*. Reading, MA: Addison-Wesley Pub. Co., 1986.

[20]   V. Zwass, "Expert system," *Britannica*, Feb. 10, 2016. https://www.britannica.com/technology/expert-system (accessed Feb. 24, 2021).

[21]   R. K. Lindsay, B. G. Buchanan, E. A. Feigenbaum, and J. Lederberg, "DENDRAL: A case study of the first expert system for scientific hypothesis formation," *Artif. Intell.*, vol. 61, no. 2, pp. 209–261, Jun. 1993, doi: 10.1016/0004-3702(93)90068-M.

[22]   L. A. Zadeh, "Fuzzy sets," *Inf. Control*, vol. 8, no. 3, pp. 338–353, Jun. 1965, doi: 10.1016/S0019-9958(65)90241-X.

[23]   J. M. Renders and J. M. Themlin, "Optimization of Fuzzy Expert Systems Using Genetic Algorithms and Neural Networks," *IEEE Trans. Fuzzy Syst.*, vol. 3, no. 3, pp. 300–312, 1995, doi: 10.1109/91.413235.

[24]   J. Straub, "Expert system gradient descent style training: Development of a defensible artificial intelligence technique," *Knowledge-Based Syst.*, p. 107275, Jul. 2021, doi: 10.1016/j.knosys.2021.107275.

[25]   S. Mitra and S. K. Pal, "Neuro-fuzzy expert systems: Relevance, features and methodologies," *IETE J. Res.*, vol. 42, no. 4–5, pp. 335–347, 1996, doi: 10.1080/03772063.1996.11415939.

[26]   J. M. McKinion and H. E. Lemmon, "Expert systems for agriculture," *Comput. Electron. Agric.*, vol. 1, no. 1, pp. 31–40, Oct. 1985, doi: 10.1016/0168-1699(85)90004-3.

[27]   E. Styvaktakis, M. H. J. Bollen, and I. Y. H. Gu, "Expert system for classification and analysis of power system events," *IEEE Trans. Power Deliv.*, vol. 17, no. 2, pp. 423–428, 2002.

[28]   O. Arsene, I. Dumitrache, and I. Mihu, "Expert system for medicine diagnosis using software agents," *Expert Syst. Appl.*, vol. 42, no. 4, pp. 1825–1834, 2015.

[29]   B. Abu-Nasser, "Medical Expert Systems Survey," *Int. J. Eng. Inf. Syst.*, vol. 1, no. 7, pp. 218–224, Sep. 2017, Accessed: Jan. 17, 2021. [Online]. Available: https://papers.ssrn.com/sol3/papers.cfm?abstract_id=3082734.

[30]   W. A. Sandham, D. J. Hamilton, A. Japp, and K. Patterson, "Neural network and neuro-fuzzy systems for improving diabetes therapy," Nov. 2002, pp. 1438–1441, doi: 10.1109/iembs.1998.747154.

[31]   E. P. Ephzibah and V. Sundarapandian, "A Neuro Fuzzy Expert System for Heart Disease Diagnosis," *Comput. Sci. Eng. An Int. J.*, vol. 2, no. 1, pp. 17–23, Feb. 2012, Accessed: Feb. 22, 2021. [Online]. Available: https://citeseerx.ist.psu.edu/viewdoc/download?doi=10.1.1.1052.1581&rep=rep1&type=pdf.

[32]   S. Das, P. K. Ghosh, and S. Kar, "Hypertension diagnosis: A comparative study using fuzzy expert system and neuro fuzzy system," 2013, doi: 10.1109/FUZZ-IEEE.2013.6622434.

[33]   B. A. Akinnuwesi, F. M. E. Uzoka, and A. O. Osamiluyi, "Neuro-Fuzzy Expert System for evaluating the performance of Distributed Software System Architecture," *Expert Syst. Appl.*, vol. 40, no. 9, pp. 3313–3327, Jul. 2013, doi: 10.1016/j.eswa.2012.12.039.

[34]   A. Chohra, A. Farah, and M. Belloucif, "Neuro-fuzzy expert system E_S_CO_V for the obstacle avoidance behavior of intelligent autonomous vehicles," *Adv. Robot.*, vol. 12, no. 6, pp. 629–649, Jan. 1997, doi: 10.1163/156855399X00045.

[35]   J. Straub, "Machine learning performance validation and training using a 'perfect' expert system," *MethodsX*, vol. 8, p. 101477, Jan. 2021, doi: 10.1016/J.MEX.2021.101477.

[36]   U. S. S. Commission, "United States Sentencing Commission: Guidelines Manual 2018," *US Sentencing Comm. Guidel. Man.*, pp. 517–522, 2018.

[37] A. Završnik, "Criminal justice, artificial intelligence systems, and human rights," *ERA Forum*, vol. 20, no. 4, pp. 567–583, 2020, doi: 10.1007/s12027-020-00602-0.

[38] M. DeMichele, P. Baumgartner, M. Wenger, K. Barrick, M. Comfort, and S. Misra, *The Public Safety Assessment: A Re-Validation and Assessment of Predictive Utility and Differential Prediction by Race and Gender in Kentucky*. 2018.

[39] "Commission Datafiles," *US Sentencing Commission*, 2021. .

[40] U. Sentencing Commission, "United States Sentencing Commission Variable Codebook for Individual Offenders Standardized Research Data Documentation for," 1999.

[41] A. V Giczy, D. Scientist, A. A. Toole, C. Economist, and N. A. Pairolero, "Identifying artificial intelligence ( AI ) invention : A novel AI patent dataset," 2021.

[42] *Conditions for patentability; novelty*. United States.

[43] *Conditions for patentability; non-obvious subject matter*. United States.

[44] "Examiner Handbook to the U.S. Patent Classification System." .

[45] "Artificial Intelligence Patent Dataset." .